\pdfoutput=1
\documentclass[11pt]{article}

\usepackage{acl}

\usepackage{times}
\usepackage{latexsym}
\usepackage{graphicx}

\usepackage[T1]{fontenc}
\usepackage[utf8]{inputenc}

\usepackage{microtype}

\usepackage{inconsolata}

\title{SPLAIN: Augmenting Cybersecurity Warnings with Reasons and Data}

\author{Vera A. Kazakova \and Jena D. Hwang \and Bonnie J. Dorr \and Yorick Wilks \\
  \{vkazakova,jhwang,bdorr,ywilks\}@ihmc.us \\ \\
  \textbf{J. Blake Gage} \and \textbf{Alex Memory} \and \textbf{Mark A. Clark}\\   
  \{john.gage,alexander.c.memory,mark.a.clark\}@leidos.com \\}

\begin{document}
\maketitle
\begin{abstract}
Effective cyber threat recognition and prevention demand comprehensible forecasting systems, as prior approaches commonly offer limited and, ultimately, unconvincing information. We introduce Simplified Plaintext Language for Actionable and Informative Narratives (SPLAIN), a natural language generator that converts warning data into user-friendly cyber threat \textbf{explanations}. SPLAIN is designed to generate clear, actionable outputs, incorporating hierarchically organized explanatory details about input data and system functionality. Given the inputs of individual sensor-induced forecasting signals and an overall warning from a fusion module, SPLAIN queries each signal for information on contributing sensors and data signals. This collected data is processed into a coherent English explanation, encompassing forecasting, sensing, and data elements for user review. SPLAIN's template-based approach ensures consistent warning structure and vocabulary. SPLAIN's hierarchical output structure allows each threat and its components to be expanded to reveal underlying explanations on demand. Our conclusions emphasize the need for designers to specify the ``how'' and ``why'' behind cyber warnings, advocate for simple structured templates in generating consistent explanations, and recognize that direct causal links in Machine Learning approaches may not always be identifiable, requiring some explanations to focus on general methodologies, such as model and training data.
\end{abstract}

\section{Introduction: Spain NLG}

Cyber-focused forecasting systems (e.g., \cite{dalton2017improving,sapienza2017early}) are becoming increasingly critical, with the growing threat and cost of cyber crime~\cite{Sobers2019,Morgan2019}. Comprehensible, reliable, and compelling forecasting systems are needed if cyber threats are to be recognized and cyber-attacks avoided.

Warnings of imminent or existing cyber threats provide a basis and an opportunity for counter-action. The likelihood of users taking effective preventative actions depends on the system's ability to accurately \textbf{predict}, but also to clearly \textbf{present} and convincingly \textbf{justify} these predictions.
Existing systems, however, provide little information beyond threat levels and confidence values. 

We design Simplified Plaintext Language for Actionable and Informative Narratives (SPLAIN), a natural language generator that converts complex warning data into human-readable \textbf{explanations}. SPLAIN generates clear and informative outputs, with actionable cyber-threat information, backed by details about input data and system functionality. 

SPLAIN’s inputs are individual sensor-induced forecasting signals coupled with an overall warning produced by a fusion module. SPLAIN queries each input to obtain information regarding contributing sensors and data signals. The collected information is combined into a single coherent narrative in English that conveys the overall warning---alongside forecasting, sensing, and data elements---to be reviewed by the user at their desired level of detail. SPLAIN employs a hierarchical template-based approach to produce warnings with consistent structure and vocabulary. Each SPLAIN output can be further expanded to reveal the underlying explanation for the overall warning.

We conclude that: (1) explainable outputs require that component designers provide the specifics of how or why (not just what) behind cyber warnings; (2) a simple set of structured templates can be developed for generating warnings and explanations with consistent sentence structure and vocabulary; and (3) direct causal links between inputs and outputs are not always identifiable within Machine Learning approaches, necessitating that some explanations describe general methodology only (e.g., model and training data).

\section{Output: Explainable Warnings}

SPLAIN is a component of a larger cyber-threat prediction system, dubbed ``Exploiting Leading Latent Indicators in Predictive Sensor Environments'' (ELLIPSE) \cite{mlg2019_14}. This system processes large quantities of web data in search of unconventional indicators of potential cyber threats, i.e. those using NLP or data foraging, as opposed to hardware-based detectors.

SPLAIN is a natural language generator that combines different levels of ELLIPSE outputs and transforms them into cyber threat predictions explained through details regarding the system's inputs and functionality. SPLAIN adheres to the following tenets:

\begin{itemize}
%\begin{AutoMultiColItemize}
    %\item actionable
    %\item accurate
    \item transparent system functionality
    \item consistent \& precise terminology
    \item hierarchically expandable structure
    \item justified assertions at all levels of explanations
%\end{AutoMultiColItemize}
\end{itemize}

\section{System: Components and Data Flow}

\begin{figure}
\includegraphics[width=1.05\linewidth]{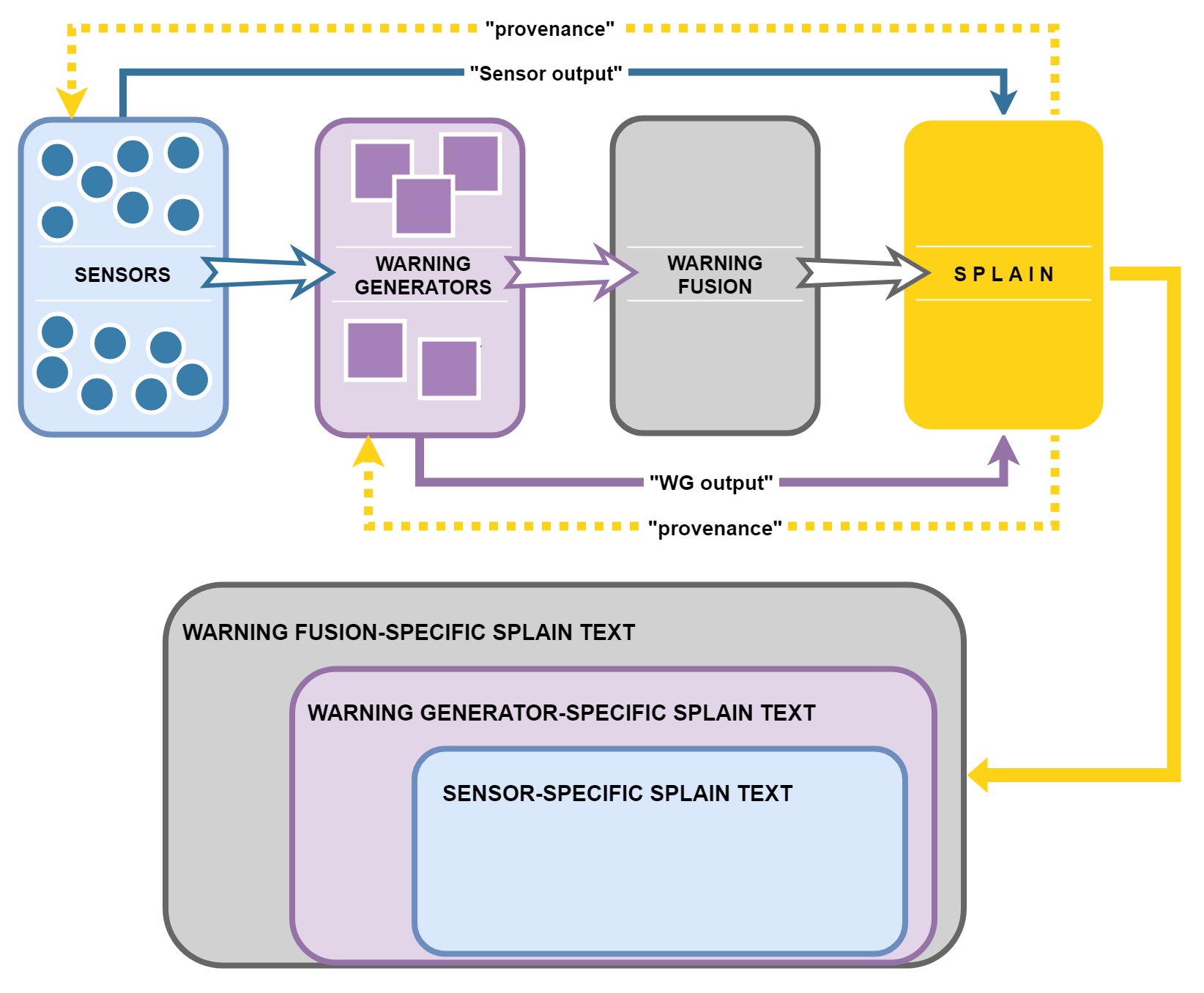}
\caption{Overall System's Data Flow}
\label{fig:overall-system-data-flow}
\end{figure}

Figure~\ref{fig:overall-system-data-flow} presents the flow of data through the ELLIPSE system, focusing on elements relevant to SPLAIN. 
Starting with \textbf{data}: files, tweets, websites feed into \textbf{sensors}, which count, score, flag, or filter these inputs. The results are then passed to \textbf{warning generators} (as well as other Sensors). These consume sensor data (standard case), raw data, model data (e.g. GloVe in "MPEMEL"), and data funneled through one-time collectors. 

Warning Generator outputs are then passed into \textbf{warning fusion}, which combines related incoming data into fused warnings. These fused outputs are passed into \textbf{SPLAIN}, which uses ID tags to navigate from fused, to warning, to sensor, to raw data, and then uses templates to generate and format the final warning along with all of its supporting data.

\section{NLG Approach: SPLAIN}

SPLAIN employs a consistent format, using precise terminology and modular structure: the design is easily modifiable if system components are altered, expanded, or replaced. A hierarchical data organization is central to the design, allowing for varying levels of explanation detail, on demand. Sensors are subdivided by type (counter, scorer, event detector, and repository). 

\begin{figure}
    \includegraphics[width=1.0\linewidth]{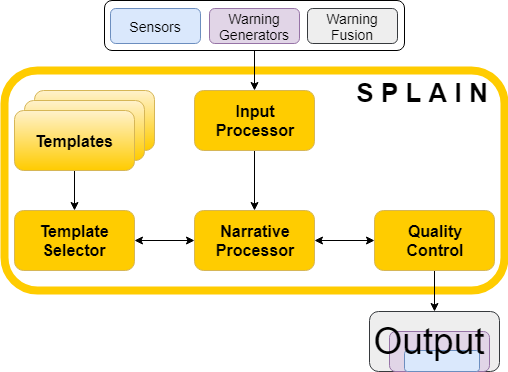}
    \caption{SPLAIN Architecture}
    \label{fig:splain-architecture}
    \end{figure}

\begin{figure}
        \includegraphics[width=1.04\linewidth]{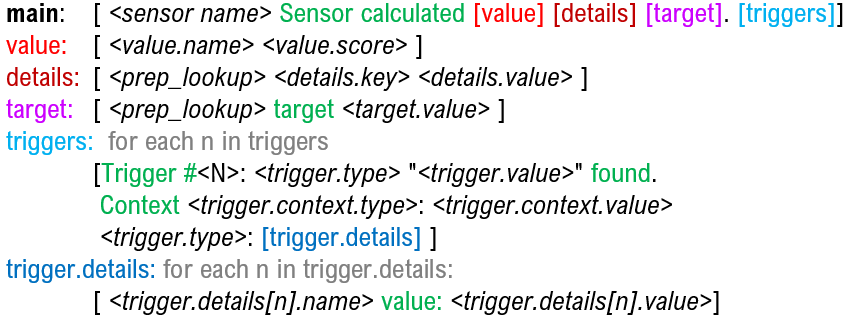}
         \caption{SPLAIN Template}
         \label{fig:spain-template}
    \end{figure}

Figure~\ref{fig:splain-architecture} presents the components of the SPLAIN design. To produce warning output, SPLAIN employs templates for scoring sensors. An example is provided in Figure~\ref{fig:spain-template}.

\section{Sample Output Warning}

ELLIPSE incorporates a variety of unconventional threat indicators, including an Outrage sensor \cite{hollingshead:2019}. An example of SPLAIN-processed output is the following generated text: 

\textit{The outrage sensor calculated an average affect value of 53\%, an average intensity value of 48.4\%, and an average outrage value of 70.91\% toward target X.} 

Additional structured representations are provided as output for this explanation, including the context for the average values presented above:\\

    \noindent
    \textbf{Trigger \#1}: term ``insanity'' found.\\
    \begin{tabular}{p{3in}}
        \small{Context tweet: "The new policies are pure insanity!"}\\
        \small{Trigger scores are: affect value of 46\%, intensity value of 55.8\%, outrage value of 71.66\%}\\
        \small{The circumplex sentiment model was used.}
    \end{tabular}

    \mbox{~~}\\
    \noindent
    \textbf{Trigger \#2}: term ``attack'' found.\\
    \begin{tabular}{p{3in}}
        \small{Context tweet: "This change is an attack on my wallet."}\\
        \small{Trigger scores are: affect value of 60\%, intensity value of 41\%, outrage value of 70.15\%}\\
        \small{The circumplex sentiment model was used.}
    \end{tabular}

These triggers would be accompanied by sources and explanations of any relevant models or approaches, such as the circumplex model~\cite{russell1980circumplex}.

\section{Conclusions}

SPLAIN employs a hierarchical template-based approach to produce expandable warnings with consistent structure and vocabulary. Each SPLAIN output can be further expanded to reveal the underlying explanation for the overall warning.  Explainable outputs require component designers to provide specifics of the ``HOW'' and ``WHY'', in addition to the ``WHAT.'' It is not always possible to identify direct causal links between the inputs and outputs of Machine Learning approaches; thus, some cyber-threat explanations must describe general methodology only (e.g., model and training data).

\section*{Acknowledgements}

This research is supported by ODNI and IARPA via the AFRL contract number FA8750-16-C-0114.

%\bibliographystyle{acl_natbib}
%\bibliography{references}

\end{document}